\documentclass[runningheads]{llncs}
\usepackage{stfloats}
\fnbelowfloat

\usepackage[T1]{fontenc}
\usepackage{graphicx}
\usepackage{xcolor}
\begin{document}
\title{
  {\small \textcolor{red}{\textbf{Accepted to Appear in the Proceedings of}}\vspace{-1.5ex}\\
  \textcolor{red}{\textbf{26th International Conference on Artificial Intelligence in Education 2025}}}\\[1ex]
  Rethinking the Potential of Multimodality in Collaborative Problem Solving Diagnosis with Large Language Models
}
\titlerunning{Rethinking Multimodality with Large Language Models}
\author{Kester Wong\inst{1}\and
Bin Wu\inst{2}\and
Sahan Bulathwela\inst{2}\and
Mutlu Cukurova\inst{1}}
\authorrunning{K. Wong et al.}
\institute{
$^{1}$ UCL Knowledge Lab, Institute of Education, University College London, UK\\
$^{2}$ AI Centre, Dept. of Computer Science, University College London, UK\\
\email{yew.wong.21@ucl.ac.uk}}
\maketitle
\begin{abstract}
Detecting collaborative and problem-solving behaviours from digital traces to interpret students' collaborative problem solving (CPS) competency is a long-term goal in the Artificial Intelligence in Education (AIEd) field. Although multimodal data and advanced models are argued to have the potential to detect complex CPS behaviours, empirical evidence on their value remains limited with some contrasting evidence. In this study, we investigated the potential of multimodal data to improve model performance in diagnosing 78 secondary school students' CPS subskills and indicators in authentic educational settings. In particular, text embeddings from verbal data and acoustic embeddings from audio data were used in a multimodal classification model for CPS diagnosis. Both unimodal and multimodal transformer-based models outperformed traditional models in detecting CPS classes. Although the inclusion of multimodality did not improve the performance of traditional unimodal models, its integration into transformer-based models demonstrated improved performance for diagnosing social-cognitive CPS classes compared to unimodal transformer-based models. Based on the results, the paper argues that multimodality and the selection of a particular modelling technique should not be taken for granted to achieve the best performance in the automated detection of every CPS subskill and indicator. Rather, their value is limited to certain types of CPS indicators, affected by the complexity of the labels, and dependent on the composition of indicators in the dataset. We conclude the paper by discussing the required nuance when considering the value of LLMs and multimodality in automated CPS diagnosis, highlighting the need for human-AI complementarity, and proposing the exploration of relevant model architectures and techniques to improve CPS diagnosis in authentic educational contexts.

\end{abstract}
\section{Introduction}
Collaborative Problem Solving (CPS) is a crucial 21st-century skill associated with communication, collaboration, and problem-solving among two or more individuals \cite{organisationforeconomicco-operationanddevelopment_2017PISAa,griffin_2012Assessment}. However, extensive international studies evaluating students' competency in CPS highlighted a general lack of proficiency among students \cite{organisationforeconomicco-operationanddevelopment_2017PISA}. There is uniform agreement about the challenges associated with the pedagogy, diagnosis, monitoring, and support of CPS in real-world contexts. For decades, Artificial Intelligence in Education (AIEd) and Computer-Supported Collaborative Learning (CSCL) communities have been working to close this gap by detecting, assessing, and supporting students' CPS development through the use of digital data and technologies (e.g., \cite{graesser2018,griffin_2017Assessing}).

Identifying indicators and aspects of CPS is the critical first step towards supporting learners and teachers. Various studies have adopted different data-driven investigations to measure CPS competency \cite{dmello_2024Improving}, describe CPS processes \cite{vrzakova_2020Focused,swiecki_2021Measuring} and predict outcomes from CPS interactions \cite{cukurova_2020Modelling,nasir_2023HMMbased} using online chat discussions \cite{An2024}, audio transcriptions \cite{stewartSayYouSay2019}, audio-acoustic features \cite{stewartMultimodalModelingCollaborative2021}, video behavioural features \cite{cukurova_2018NISPI}, physiological features \cite{Giannakos2019}, and user logs \cite{rojas_2021Assessing} of students. 

Despite the demonstrated benefits of integrating multimodal data for triangulation and better contextualization of information \cite{muMultimodalDataFusion2020}, most previous work relies on manual coding of CPS indicators. This significantly reduces research efficiency and introduces inconsistencies even with trained human coders. Hence, attempts have also been made to detect CPS indicators automatically using analytics, rule-based algorithms, and machine learning (ML) techniques \cite{Adams2015,Khan2017,stewartMultimodalModelingCollaborative2021}. However, there is limited empirical work on the potential of recent AI developments, such as pre-trained large language models (LLMs), to advance such efforts further. In particular, most studies, if not all, which leveraged LLMs to detect aspects of CPS have only used transcription from verbal data (i.e., unimodal LLM) \cite{zhao_2024automated,dmello_2024Improving,pugh_2021Say}. The potential inclusion of multimodality in LLMs for CPS indicator diagnosis remains largely unexplored. Hence, this study uses information-rich representations (i.e., text and audio embeddings of audio data) of multimodal data instead of interpretable feature representations for CPS diagnosis.

\section{Related Work}
Previous research that applied AI techniques to detect CPS indicators was reviewed, focusing on work involving multimodal data to improve model performance for automated detection of CPS behaviours, subskills, or competency.

\subsection{Collaborative Problem Solving Subskills and Indicators} \label{CPSfacet}
Various definitions of CPS have been proposed in the literature. While some may view it as a group of individuals coming together to work on certain problem-solving activities \cite{oneil_2010ComputerBased}, others elaborated on the interactions that take place with the characterisation of the group as a collective entity \cite{hesse_2015Framework}. Regardless, most researchers agree that CPS is a multilevel process involving the perspectives of both individuals and groups. It should consider the cognitive domain knowledge that is activated during CPS alongside the social aspects of interacting with others to progress towards a shared understanding \cite{luckin_2017Solved}. In addition, these processes are highly dynamic, with group interactions and individual internalisation of cognitive and social components having mutual effects on each other. As such, it is difficult to ascertain the quality of CPS by unguided observations. However, certain mechanisms proposed in the learning sciences literature may be adopted as a lens to understand how certain observable behaviours during CPS may influence certain dimensions, such as cognition. These observable behaviours are frequently termed as "\textit{indicators}", and a collection of different associated indicators are reflective of latent variables that are termed as "\textit{facets}" when referring to CPS competency \cite{sunGeneralizedCompetencyModel2020} or "\textit{subskills}" when describing CPS processes \cite{taylor_2024Quantifying}.

\subsection {AI Approaches to Detect CPS Indicators, Facets, or Subskills}
AI techniques have been applied in various studies to detect nonverbal and verbal aspects of CPS. For example, Zhou et al. \cite{Zhou2024} used speaker identification and diarization to detect nonverbal speech behaviours, and processed data of facial features with convolutional neural networks (CNN) to detect gaze behaviours and different types of group interactions among peers. Spikol et al. \cite{Spikol2018} collected data on students' physical movements, facial features, audio speech, and interactions with a sentiment button in small groups. They used features from these data to construct a deep neural network (DNN) to predict group performance. However, most of these studies relied heavily on manual labelling of CPS indicators. The work by D'mello et al. \cite{dmello_2024Improving} was one of the few and most recent works that leveraged a Natural Language Processing (NLP) technique to automatically detect the probability that a given utterance from automated transcription was a label of three CPS facets to generate a score on students' CPS competency. However, the real-world implementations of these models in noisy classroom environments struggled to generate reliable and accurate indicator detections \cite{pugh_2021Say}. This study aims to advance this line of work by investigating the potential of recent LLMs and multimodal data to automate the diagnosis of CPS indicators.

\subsection{Diagnosis of CPS from Verbal Data with NLP and the Value of Multimodality}
In the learning analytics literature, a relatively small number of studies sought to use ML to train models on human-labelled datasets to \textit{automatically} detect CPS indicators and facets from verbal data. Relatively simpler NLP approaches such as Bag-of-Words (BoW) models or Linguistic Inquiry Word Count (LIWC) have been used to represent transcribed utterances from a lexical level \cite{stewartSayYouSay2019}. In contrast, more recent and advanced models like the Bidirectional Encoder Representations from Transformers (BERT) \cite{devlin_2019BERT}, which are capable of representing nuances, dependencies, and relationships between words in the transcribed utterance by leveraging on deep neural networks, demonstrated higher performance for the detection of CPS indicators \cite{pugh_2021Say}. Using pre-trained transformer-based models have shown improvements in many AIEd tasks \cite{bulathwela2023scalable,li2025novel}.

However, it is argued that building models on unimodal data, such as verbal speech transcription to detect CPS with ML techniques, limits the ability of the model to capture the complexity of interactions during CPS \cite{Cukurova2020_promise}. Previous research demonstrated that the fusion of multimodal data could indeed produce significantly better prediction models for complex learning constructs (e.g., \cite{Giannakos2019,cukurova_2020Modelling}). Yet, other studies have suggested limitations of incorporating multimodality to detect CPS automatically. For instance, Stewart et al. \cite{stewartMultimodalModelingCollaborative2021} investigated how multiple modalities of data could improve the detection of three CPS facets. These modalities involved transcription extracted using Automatic Speech Recognition (ASR) software (i.e., IBM Watson Speech to Text service) from audio files, log data of change in on-screen interaction, facial features and motion of video recordings extracted by a computer vision software (i.e., Emotient) \cite{Littlewort2011}, and acoustic-prosodic features of audio recordings extracted by \texttt{openSMILE} \cite{Eyben2013}. It was found that adding facial expressions and acoustic-prosodic features to the model trained on transcription features and task context actually \textit{decreased classification accuracy} for all three CPS facets. This finding may be due to indicators used in the CPS framework which were specifically centred on conversation features (e.g., "\textit{initiates off-topic conversation}") and task-related behaviours (e.g., "\textit{proposes specific solutions}") (see \cite{sunGeneralizedCompetencyModel2020} for all the indicators). 

As part of the ongoing discussions in the field regarding the generalisability of multimodal learning analytics (e.g. \cite{chejara2023build}), it is unclear to what extent such findings on the potential or detrimental effects of multimodality on the model performance of detecting CPS indicators can be generalised. With various studies having investigated the use of audio channels for emotion recognition \cite{anagnostopoulos_2015Features}, it remains to be explored if certain CPS indicators involving both cognitive and affective constructs (e.g., perspective taking \cite{hesse_2015Framework}) may see improved detection by multimodal models. A more detailed analysis of the specific detection of CPS indicators is needed to understand the potential value and limitations of multimodal data, particularly in light of work utilising more advanced AI techniques such as LLMs to detect CPS indicators and facets automatically. 

\subsection{Current Study}
This study explores the use of multimodal audio speech data involving verbal text content (i.e., transcription) and audio features (i.e., acoustic-prosodic) for CPS diagnosis. The approaches and methods used in this study are key contributions to the literature as they seek to propose meaningful ways of harnessing multimodality in advanced NLP techniques for automated diagnosis of CPS. There are two research questions posed in this study:
\newline \textbf{RQ1}: How does the performance of a transformer-based model compare to a traditional machine learning model for detecting CPS classes using audio data?\newline
\textbf{RQ2}: How does the performance of models detecting CPS classes differ when using multimodal audio data involving verbal text content and audio features?\newline

\section{Methodology}
The approaches used when developing the models, datasets and the experiments to answer the research questions are outlined in this section.

\subsection{Models}
\subsubsection{Traditional machine learning models using extracted statistical features:}\label{model_rf}
This study involved traditional machine learning models using extracted statistical features as the baseline to answer RQ1 and RQ2. In particular, Random Forest (RF) classifier was applied on a Term Frequency-Inverse Document Frequency (TF-IDF) representation. Previous research indicates that RF classifiers can outperform or achieve comparable performance to other models (e.g., Na\"ive Bayes, Support Vector Machine) for CPS facet classification \cite{stewartMultimodalModelingCollaborative2021}. In addition, TF-IDF representation of verbal text content captures the difference of text in the whole data and has been a useful baseline for various NLP tasks \cite{kumar_2021Ensembling,zhang_2008TFIDF}. The TF-IDF vectorisation produced sparse vector representations of the text after removing common English stop words. Hyperparameters were optimised via three-fold cross-validation split where the folds were sampled with stratification. A grid search was performed to identify the optimal hyperparameter combination using average accuracy across the three splits as its selection criterion.

\subsubsection{Transformer-based models using verbal text content:}\label{model_bert}
The BERT model is a powerful transformer-based model used to generate text embeddings. Various recent studies have explored the use of it for automated detection of indicators related to CPS (e.g., \cite{zhao_2024automated,dmello_2024Improving,pugh_2022SpeechBased}). We use it as the baseline to answer RQ2. First, feature extraction is performed by processing the transcribed utterances using \texttt{BertTokenizer} that uses the \textit{WordPiece} algorithm.

Since BERT model was pre-trained on an extensive general-purpose text corpus, there was a lack of domain-specific knowledge. Fine-tuning, a widely-used technique, was used to close this domain knowledge gap \cite{tejani_2022Performance}. Specifically, we fine-tuned the \texttt{bert-base-uncased}\footnote{\url{https://huggingface.co/google-bert/bert-base-uncased}} on utterances in the training data in each of the datasets. The training data was randomly split into 80\% for model training and 20\% for model validation, using a fixed random seed for reproducibility. Fine-tuning was conducted over multiple iterations, utilizing the \texttt{AdamW} optimiser with a learning rate of $2 \times 10^{-5}$ and an epsilon value of $1 \times 10^{-8}$ for numerical stability. The final model used for performance evaluation on the testing data was selected based on the epoch that achieved an optimal balance between predictive performance (validation F1 score) and generalizability (validation loss).

\subsubsection{Multimodal models using verbal text content and audio features:}
Two multimodal models were trained in this study. One of the models is an extension of the RF TF-IDF model, described in section \ref{model_rf}, where 11 audio features (previously selected by \cite{stewartMultimodalModelingCollaborative2021}) extracted using \texttt{openSMILE} were concatenated with the TF-IDF vector (RF TF-IDF+A). \texttt{OpenSMILE} was configured with a $1$ ms frame step to compute low-level descriptors (LLDs) from the \texttt{GeMAPSv01b} feature set. The audio was resampled to a uniform sampling rate of $16$ kHz before extraction to ensure consistent processing of all audio files for reliable comparison between extractions. This model is used as a baseline to answer RQ1.

The second model is a multimodal variant of the BERT model, Audio-Assisted BERT (AudiBERT). Toto et al. \cite{toto_2021AudiBERT} proposed the design of AudiBERT involving an audio-textual dual self-attentive framework that captures interdependence between different parts of speech in audio and textual form. They evaluated AudiBERT with various text and audio representation models and showed improved performance compared to unimodal models. We apply the AudiBERT infrastructure using BERT as the text encoding model (same parameters as in section \ref{model_bert}) and Wav2Vec2.0 as the audio encoding model. Wav2Vec2.0\footnote{https://huggingface.co/facebook/wav2vec2-base-960h} produces CNN embeddings from audio segments, which are processed to produce transformer embeddings. Transfer learning was applied by initialising the AudiBERT text encoder with pre-trained weights of the encoding model from the selected epoch of the BERT model. Instead of training from scratch, this technique reduces training iterations required to achieve optimal convergence while exploiting prior knowledge learnt from pre-training.

Both BERT and Wav2Vec2.0 embeddings were obtained and separately processed as a sequence in their respective Bidirectional Long Short-Term Memory (BiLSTM) network to obtain a layer of hidden state vectors. A self-attention mechanism is then applied to this layer to obtain audio and text embeddings. These two embeddings are finally fused together by concatenating them together to represent a particular utterance (see \cite{toto_2021AudiBERT} for algorithm). After that, the same fine-tuning process (training validation split and cross-entropy loss function) and selection method applied to the BERT model in section \ref{model_bert} was undertaken\footnote{https://github.com/keswong/AIED25-AudiBERT.git}.

\subsection{Participants and Context}
The study involved 78 Secondary school students (aged 14 - 15 years) from a public school. Participants were recruited from mathematics classes of similar academic ability. Before the CPS task, a pre-test was administered to assess their ability to solve a similar problem as the CPS task. Participants from the same class were grouped in the same triad to ensure familiarity with each other \cite{Gruenfeld19961}. Statistical analysis was done to ensure that no significant differences in pre-test scores were observed between triads since CPS relies on domain knowledge \cite{MayerMerlin1996}. The aim was to form symmetrical groups based on knowledge and status \cite{Dillenbourg1999}. The complete dataset involved 26 triads (18 mixed-sex, 4 all-female, 4 all-male). 

Students were given 50 minutes to work in their assigned triad on a CPS task consisting of one mathematics problem-solving question that involved the application of two concepts: Pythagoras Theorem and quadratic equations. A curriculum expert vetted the question for correctness, and participants' teachers deemed it to be of suitable difficulty and relevance to the curriculum. Since students collaborated without teacher assistance, scripting of collaboration and the problem-solving process was structured into the task to support them \cite{Fischer2007}. 

\subsection{Data Processing}
Data was collected using video conferencing software, where students worked remotely together and had their individual audio files recorded to eliminate the need for speaker diarization. Transcriptions were generated from these audio files using the ASR system \texttt{ins8.ai}, a speech-to-text model trained on a mix of open source and localised audio datasets that were contextualised to the colloquial speech spoken by the students. The first author manually corrected inaccurate transcriptions. The Word Error Rate (WER) for each utterance computed using \texttt{Jiwer} \cite{jiwer2019} yielded an average WER of $0.553$ across 8692 utterances.

\subsection{Human Coding of Collaborative Problem Solving Indicators}
The theoretical framework used in this study (see appendix) is adapted from Taylor et al. \cite{taylor_2024Quantifying} as the proposed indicators involved problem-solving practices. Taylor et al. applied the framework and coded at the subskills level. However, since coding at the indicator level was done in this study, subtle variations in indicator descriptions made it challenging to code distinctly. For example, Taylor et al. distinguished "\textit{Naming difficulties and limitations that obstruct the group from addressing the problem}" and "\textit{Identifying the need for more information related to the problem}" as two distinct indicators. Yet, articulating difficulties implicitly involves identifying the needed information (e.g., difficulty in using a formula to solve an equation implies the identified need for information about the formula). Hence, the 47 indicators in the framework by Taylor et al. were narrowed to 42 indicators. The subskills were then realigned to key facets of CPS discussed in Section \ref{CPSfacet} and aspects of the problem solving process, refining the 19 subskills in the framework by Taylor et al. to 10 critical subskills (i.e., \textit{SS1} - \textit{SS10}). Since scripting was a crucial aspect of the CPS task and process, five indicators on the dimension of scripting were also included in the framework and classified under two subskills (i.e., \textit{SC1} \& \textit{SC2}). These indicators in the social-cognitive dimensions were concurrently coded with 10 other indicators across three affect states (i.e., \textit{AS1} - \textit{AS3}) on the affective dimension to capture the emotional exchanges among students during their interaction.

One researcher and one school teacher, both with at least ten years of teaching experience, double-coded a total of 1524 utterances (17.5\% of the total transcription utterances) and achieved a Cohen's kappa interrater reliability of 0.847. Coders were then randomly assigned triads to code the remaining utterances. This produced 2109 utterances coded on 34 problem-solving indicators, 2449 utterances on all scripting indicators, and 1528 utterances on all affect indicators.

Named Entity Recognition (NER) was applied using the pre-trained NER model from spaCy \texttt{en\_core\_web\_sm} module package to reduce the noise and improve model performance during classification \cite{mohit_2014Named}. Human correction to inaccurate NER annotation was made by editing 1342 utterances (15.4\% of the dataset), of which 651 were modifications on the edited utterances by the NER model. These annotations involved 11 different masks associated with names, web-based resources, locations, entertainment and devices used in the CPS task.

Utterances with multiple coded indicators were split into individual instances, with each instance tagged to a single indicator. This ensures that the classification does not capture relationships between multiple classes, as the framework does not contain such dependencies across different indicators.

\subsection{Dataset}
The dataset is partitioned on the social-cognitive and affective dimensions for models to perform classification of subskills and states within each dimension independently. Two utterances coded on subskills with less than ten instances were removed. Table \ref{tab:utterancecoded} shows the breakdown of the number of utterances in each of the datasets after performing 80:20 stratified train-test split.

\begin{table}[!hb]
\small
\renewcommand{\arraystretch}{1.2} 
\setlength{\tabcolsep}{2.5pt} 
\centering
\begin{tabular}{c|cccccccccc|ccc}
\hline
Dimension     & \multicolumn{10}{c|}{Social-cognitive}                      & \multicolumn{3}{c}{Affective} \\
Dataset       & SS1 & SS2  & SS3 & SS4 & SS5 & SS6 & SS7 & SS8 & SC1  & SC2 & AS1      & AS2      & AS3     \\ \hline
Total data    & 215 & 1223 & 349 & 27  & 94  & 126 & 22  & 51  & 1590 & 859 & 569      & 920      & 39      \\
Training data & 172 & 978  & 279 & 21  & 75  & 101 & 18  & 41  & 1272 & 687 & 455      & 736      & 31      \\
Testing data  & 43  & 245  & 70  & 6   & 19  & 25  & 4   & 10  & 318  & 172 & 114      & 184      & 8       \\ \hline      
\end{tabular}
\caption{Dataset of subskills and states for training and testing.}
\label{tab:utterancecoded}
\end{table}

\subsection{Experimental Design and Evaluation Metrics}
TF-IDF with BERT, and TF-IDF+A with AudiBERT were respectively compared to answer RQ1 relating to investigation on transformer-based models over traditional machine learning models. In addition, comparisons of TF-IDF with TF-IDF+A, and BERT with AudiBERT were performed to investigate the potential of multimodality and answer RQ2. These experiments were performed using the social-cognitive dataset and the affective dataset separately.

These experiments were first evaluated by comparing model performance using the weighted precision, recall, and F1 score. These were calculated using labels coded by human experts (i.e. true labels) and predicted labels by the different models. Subsequently, confusion matrices were used to analyse the distribution of predictions by the model. The values in the confusion matrices were normalised by dividing each value by the total number of true labels in its respective class. The row-normalised confusion matrix was represented with a heatmap to visualise the distribution of labels across the different classes.

\section{Results}
\begin{figure}[!ht]
    \centering
    \includegraphics[width=0.935\linewidth]{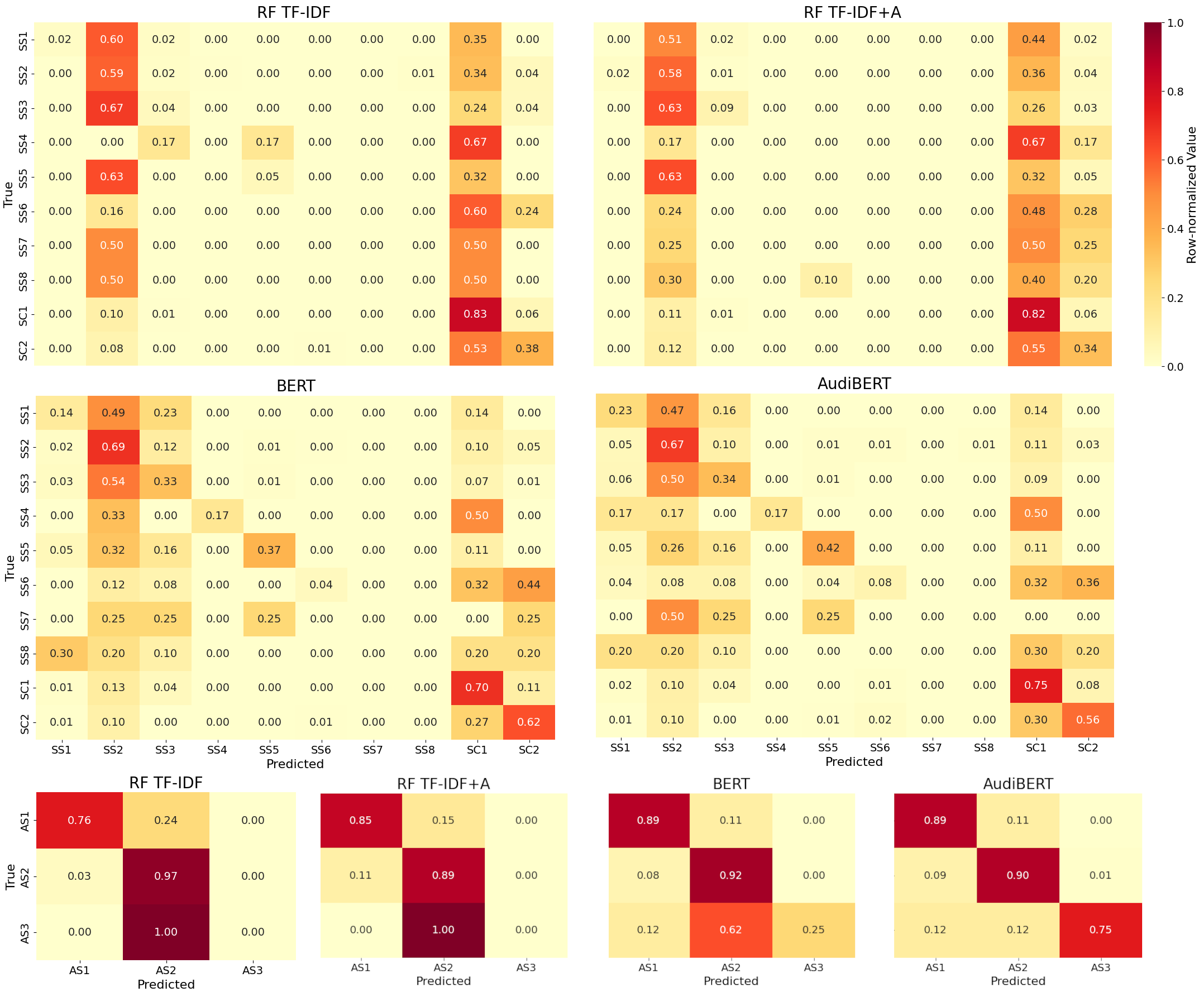}
    \caption{Row-normalized confusion matrix with values in each cell showing the proportion of instances from the true class that were predicted as a particular class in the social-cognitive (rows 1 and 2) and in the affective dimension (row 3).}
    \label{fig:CM}
\end{figure}

The performance metrics and row-normalised confusion matrices of the respective models are provided below in Table \ref{tab:performance_table} and Figure \ref{fig:CM} respectively.

\begin{table}[!hb]
\small
\setlength{\tabcolsep}{8pt} 
\centering
\begin{tabular}{c|cccc|cccc}
\hline
\multicolumn{1}{r|}{Dimension}   & \multicolumn{4}{c|}{Social-Cognitive} & \multicolumn{4}{c}{Affective} \\
\multicolumn{1}{l|}{Model}       & Acc. & Prec. & Rec. & F1 & Acc. & Prec. & Rec. & F1 \\ \hline
RF TF-IDF   & .524& .514     & .524  & .468 & .866& .851& .866& .852\\
RF TF-IDF+A & .512& .459     & .512  & .455 & .853& .831& .853& .842\\
BERT        & \emph{.589}&\emph{.573}     & \emph{.589}  & \emph{.573} & \textbf{.892}& \textbf{.894}  & \textbf{.892}  & \emph{.887}  \\
AudiBERT    & \textbf{.598}& \textbf{.587}     & \textbf{.598}  & \textbf{.587} & \textit{.889}& \emph{.890}  & \emph{.889}  & \textbf{.889}  \\ \hline
\end{tabular}
\caption{Comparison of predictive Accuracy (Acc.), Precision (Prec.), Recall (Rec.) and F1-Score (F1) between models on testing data. The best and second best performances are indicated in \textbf{bold} and \emph{italic} faces respectively. The classification reports of the models are provided in the digital appendix.}
\label{tab:performance_table}
\end{table}

\subsection{Performance Differences between Models (RQ1)}

In terms of detecting CPS classes using audio data, transformer-based models demonstrated better performance on weighted precision, recall, and F1 scores over traditional models for classification on classes in both social-cognitive and affective dimensions (refer to Table \ref{tab:performance_table}). Overall, there was a higher performance in detecting the small number of classes in the affective dimension (i.e., metric values > 0.8) than in the social-cognitive dimension (i.e., metric values < 0.6).

It was observed that the traditional machine learning models tend to classify the text into several specific classes (i.e., \textit{SS2}, \textit{SC1} and \textit{SC2} for social-cognitive subskills, and \textit{AS2} for affect states) that had a large number of utterances among the classes in the training dataset (refer to Figure \ref{fig:CM}). Transformer-based models alleviate this issue by providing a wider range of true positive classifications (i.e., \textit{SS1}, \textit{SS2}, \textit{SS3}, \textit{SS4}, \textit{SS5}, \textit{SC1} and \textit{SC2}), enabling prediction of classes that are sparse in the dataset (i.e., \textit{SS4}, \textit{SS5}, \textit{AS3}) to improve overall performance.

\subsection{Performance Differences between Modalities (RQ2)}
It is observed in Table \ref{tab:performance_table} that the RF TF-IDF+A model had lower performance than the TF-IDF model. Including audio features could have introduced variability and noise to the dataset, making it difficult for the model to distinguish relevant patterns. Yet, this effect seems to be minimal for transformer-based models. AudiBERT is observed to outperform BERT on F1 scores for classifications in both social-cognitive and affective dimensions. However, BERT was observed to perform better in accuracy, precision and recall compared to AudiBERT for classifications in the affective dimension. Figure \ref{fig:CM} showed that unimodal and multimodal models made predictions predominantly on the same classes, with few or no predictions on "\textit{SS6}", "\textit{SS7}", or "\textit{SS8}" which were underrepresented in the dataset (refer to Table \ref{tab:utterancecoded}). However, AudiBERT was able to improve the classification of "\textit{AS3}" compared to BERT, despite having significantly fewer training data than the other classes in the affective dimension.

\section{Discussion}
This study demonstrated the potential use of audio data to enable multimodality in transformer-based models to improve the overall performance of diagnosis of CPS classes in the social-cognitive dimension. The confusion matrix showed that Audibert was able to improve the classification of "\textit{SS1: Sense-making}", "\textit{SS3: Formulating a solution}", "\textit{SS5: Reaching a solution}", "\textit{SS6: Maintaining roles and responsibilities}" and "\textit{SC1: Using scripting}". Practioners may consider using multimodal audio data to improve the automated diagnosis of these particular CPS classes that may be highly nuanced and difficult for unimodal transformer models to classify accurately.

Similar improvement was not present for the diagnosis of affective CPS classes although existing studies have investigated the empirical detection of acoustic differences between emotions \cite{Scherer2003} and the use of acoustic features to reflect physiological stress \cite{Kappen2024}. This disparity may stem from the methodology used, where human-coded CPS labels were based on semantic rather than emotional analysis. This consequently can make the labelling more nuanced and aligned with text-based representations. Hence, word representations in the utterances corresponding to each class had the potential for detecting CPS indicators related to the affect dimension. Previous studies on emotion and language also highlighted the role of semantics in emotion classification \cite{Johnson-Laird1989}, which might be the reason behind observed results. Improvements by multimodality are couched on the choice of the model architecture, and the association between the multimodal features and labelled classes. Hence, one possible approach for tackling complex diagnostic tasks such as CPS would be to use ensemble learning methods with different classification models that perform well in diagnosing specific CPS indicators. The predictions of these base classifiers can be combined using a meta learner (i.e. stacking) among many other aggregation methods, to produce a final prediction of CPS indicators  \cite{ganaie_2022Ensemble}.

In addition, the results suggest that complex models may not necessarily improve the classification produced by simpler models with already robust performance. As discussed by Suraworachet et al. \cite{Suraworachet2024}), \textit{"... which method should be used predominantly depends on the use case, goals ... as well as multiple social and ethical considerations"}. Although there is much interest and hype surrounding the potential of LLMs, the results show that the generic argument of prioritizing one modelling technique over another to detect CPS has its limitations. It is important to consider different modelling approaches based on the distribution of classes in the specific datasets and CPS diagnosis tasks. In this study, the dataset was partitioned along social-cognitive and affective dimensions, and CPS diagnosis was performed independently in each dimension. 
A different approach would be to involve learning social-cognitive and affective classes together for neural models to share represents between the two task groups and enhance task performance (multi-task learning). Multi-task direction is further justified since the coding of social-cognitive and affective labels was performed concurrently by the expert coders \cite{Jin2020Multi}.

Furthermore, this study identified several well-detected classes involving three subskills and three affective states. These subskill classes were "\textit{SS2: Building shared understanding}", "\textit{SC1: Using scripting}", "\textit{SC2: Regulating scripting}", "\textit{A1: Neutral affect state}", "\textit{A2: Negative affect state}", and "\textit{A3: Positive affect state}". The others were too complex for even the best models to detect. This highlights the potential of designing systems whereby classes that are semantically clearer than other classes are leveraging AI models for automated labelling and leave complex labelling decisions to human experts, enhancing productivity in human CPS diagnosis tasks through human-AI complementarity \cite{Cukurova2024}.

\subsection{Limitations and Future Work}
The small sample size for particular classes reduced statistical power for interpretation. However, increased student sampling may not necessarily yield significant increments for certain classes since the lack of holistic CPS proficiency among students could still lead to data scarcity on certain CPS indicators. Data augmentation techniques could be used to provide more examples for model training and improve model performance \cite{Shorten2021}.

Furthermore, there may be a ceiling effect on the performance of large language models due to their inherent architecture. While various approaches (e.g., transfer learning, including multimodality) may be taken to improve classification, there may be diminishing returns to model improvements \cite{luo_2024Has}. 
Using BERT initialisations that are specifically pre-trained on mathematic data might also lead to improvements as shown in prior works in education \cite{bulathwela2023scalable}.

While the results are promising, there is more mileage to cover in terms of running more rigorous experiments, involving multifold cross-validation, that support statistical analysis of performance across several different models \cite{rainio_2024Evaluation}. Fine-grained research questions such as what specific type of improvements differentiate BERT from AudiBERT, and the correlation between training data size and class performance need to be explored deeper to get a good understanding of the impact of incorporating multimodal features in CPS detection models. Trying to determine feature importance-based explanations will also help scrutinise the inner workings of transformer-based models. 

Models involving speech recognition may contain potential biases due to gender and sociocultural differences present in speech patterns and language use \cite{Liu2022Towards,tatman-2017-gender}. Hence, the models presented here may result in systematic inaccuracies in CPS diagnosis when directly applied to a different dataset. The generalisability of the findings may vary across different semantic contexts of audio data. Similar studies on using models for CPS diagnosis should adopt a similar methodology to obtain a fine-tuned model for their specified use case. Further investigation should also be conducted to explain what tokens the model draws attention from to determine the classes. This contributes toward model explainability \cite{10700713} and could potentially improve awareness of inherent model biases \cite{CHUAN2024102135}.

\section{Conclusion}
Diagnosing relevant indicators for supporting teaching and learning practice in CPS is challenging. The use of recent AI models provides the potential to automate the coding process. However, it is constrained by the affordances of different models, relevant inclusion of different modalities, and types of data for training these models to detect CPS. Although the general potential of multimodality with LLMs was observed for certain aspects of CPS diagnosis in this study, these approaches should not be regarded as a panacea for automatically diagnosing all CPS indicators, facets, and competency due to numerous other factors influencing the performance of models. These results accentuate the need for more nuanced discussions about the value of any particular modelling approach (including multimodality and LLMs) on the diagnosis of complex learning processes like CPS in research and practice. Here, we direct the thinking of such discussions, in AIEd and beyond, towards the ideas of ensemble model architectures and human-AI complementarity frameworks to advance research focusing on the complex task of diagnosing CPS, consequently providing greater opportunities for supporting CPS practice and research.

\begin{credits}
\subsubsection{\discintname}
The authors have no competing interests to declare that are relevant to the content of this article.
\end{credits}

\subsubsection{\small Appendix.}
\small
CPS framework and classification report of models provided below.\\
\url{https://osf.io/u4nvb/?view\_only=3b02f769ea4746a0bbbae514ab1acb8e}

\bibliographystyle{splncs04}

\end{document}